\documentclass{article}
\usepackage[utf8]{inputenc} 
\usepackage[T1]{fontenc}    
\usepackage{hyperref}       
\usepackage{url}            
\usepackage{booktabs}       
\usepackage{fullpage}

\usepackage{amsfonts}       
\usepackage{nicefrac}       
\usepackage{microtype}      
\usepackage{array}          
\usepackage{tabularx}       
\usepackage{enumitem}
\usepackage{mathtools}
\usepackage{amsmath}
\usepackage{bm}
\usepackage{amsthm}
\usepackage{graphicx}
\usepackage{subcaption}
\usepackage{multirow}
\usepackage[binary-units=true]{siunitx}
\usepackage{makecell}
\usepackage{setspace}
\usepackage{todonotes}
\usepackage{xcolor}
\RequirePackage{algorithm}
\RequirePackage{algorithmic}
\RequirePackage{natbib}

\makeatletter
  \def\title@font{\Large}
  \let\ltx@maketitle\@maketitle
  \def\@maketitle{\bgroup%
    \let\ltx@title\@title%
    \def\@title{\resizebox{\textwidth}{!}{%
      \mbox{\title@font\ltx@title}%
    }}%
    \ltx@maketitle%
  \egroup}
\makeatother

\DeclareSIUnit\px{px}
\pdfoutput=1

\newtheorem{prop}{Proposition}

\newtheorem{rmq}{Remark}

\newtheorem{corollary}{Corollary}

\newtheorem{definition}{Definition}

\newtheorem*{example*}{Example}
\newtheorem*{prop*}{Proposition}
\newtheorem*{thm*}{Theorem}
\newtheorem*{prv*}{Proof}
\newtheorem*{rmq*}{Remark}

\newcommand{\ie}{\textit{i.e.}}
\newcommand{\st}{s.t. }
\newcommand{\Lip}{\mathrm{Lip}}

\newcommand{\RR}{\mathbb{R}}
\newcommand{\inputspace}{\mathcal{X}}
\newcommand{\labelspace}{\mathcal{Y}}

\DeclareMathOperator*{\argminB}{argmin}   
  
\DeclareMathOperator*{\argmaxB}{argmax}

\makeatletter
\def\blfootnote{\gdef\@thefnmark{}\@footnotetext}
\makeatother

\title{A Dynamical System Perspective for Lipschitz Neural Networks}

\author{
    Laurent Meunier$^{*1,2}$, Blaise Delattre$^{*2}$, Alexandre Araujo$^{*3}$, Alexandre Allauzen$^{2,4}$ \\[0.3cm]
    $^{1}$ Facebook AI Research \\
    $^{2}$ LAMSADE, Université Paris-Dauphine, CNRS, PSL University \\
    $^{3}$ INRIA, Ecole Normale Supérieure, CNRS, PSL University \\
    $^{4}$ ESPCI Paris
}

\date{}

\begin{document}
\maketitle

\blfootnote{$^*$ Equal contributions}

\begin{abstract}
The Lipschitz constant of neural networks has been established as a key quantity to enforce the robustness  to adversarial examples. In this paper, we tackle the problem of building $1$-Lipschitz Neural Networks. By studying Residual Networks from a continuous time dynamical system perspective, we provide a generic method to build $1$-Lipschitz Neural Networks and show that some previous approaches are special cases of this framework. Then, we extend this reasoning and show that ResNet flows derived from convex potentials define $1$-Lipschitz transformations, that lead us to define the {\em Convex Potential Layer} (CPL). A comprehensive set of experiments on several datasets demonstrates the scalability of our architecture and the benefits as an $\ell_2$-provable defense against adversarial examples.
\end{abstract}

\section{Introduction}

Modern neural networks have been known to be sensible against small, imperceptible and adversarially-chosen perturbations of their inputs~\citep{biggio2013evasion,szegedy2014intriguing}.
This vulnerability  has become a major issue as more and more neural networks have been deployed into production applications.
Over the past decade, the research progress plays out like a cat-and-mouse game between the development of more and more powerful attacks~\citep{goodfellow2014explaining,kurakin2016adversarial,carlini2017adversarial,croce2020reliable} and the design of empirical defense mechanisms~\citep{madry2017towards,moosavi2019robustness,cohen2019certified}.
Finishing the game calls for certified adversarial robustness~\citep{raghunathan2018certified,wong2018scaling}.
While recent work devised defenses with theoretical guarantees against adversarial perturbations, they share the same limitation, \ie, the tradeoffs between expressivity and robustness, and between scalability and accuracy.

A natural approach to provide robustness guarantees on a classifier is to enforce Lipschitzness properties. 
To achieve such properties, researchers mainly focused on two different kinds of approaches.
The first one is based on randomization~\citep{lecuyer2018certified,cohen2019certified,pinot2019theoretical} and consists in convolving the input with with a predefined probability distribution.
While this approach offers some level of scalability (\ie, currently the only certified defense on the ImageNet dataset), it suffers from significant impossibility results~\cite{yang2020randomized}.
A second approach consists in building $1$-Lipschitz layers using specific linear transform~\citep{cisse2017parseval,li2019preventing,anil2019sorting,trockman2021orthogonalizing,skew2021sahil,li2019preventing,singla2021householder}.
Knowing the Lipschitz constant of the network, it is then possible to compute a certification radius around any points. 

A large line of work explored the interpretation of residual neural networks \cite{he2016deep} as a parameter estimation problem of nonlinear dynamical systems~\citep{haber2017stable,e17Proposal,lu18beyond}.
Reconsidering the ResNet architecture as an Euler discretization of a continuous dynamical system yields to the trend around Neural Ordinary Differential Equation~\citep{chen2018neural}.
For instance, in the seminal work of~\citet{haber2017stable}, the continuous formulation offers more flexibility to investigate the stability of neural networks during inference, knowing that the discretization will be then implemented by the architecture design.
The notion of stability, in our context, quantifies how a small perturbation on the initial value impacts the trajectories of the dynamical system. 

From this continuous and dynamical interpretation, we  analyze the Lipschitzness property of Neural Networks. We then study the discretization schemes that can preserve the Lipschitzness properties. With this point of view, we can readily recover several previous methods that build 1-Lipschitz neural networks~\citep{trockman2021orthogonalizing,skew2021sahil}.
Therefore, the dynamical system perspective offers a general and flexible framework to build Lipschitz Neural Networks facilitating the discovery of new approaches.
In this vein, we introduce convex potentials in the design of the Residual Network flow and show that this choice of parametrization yields to by-design $1$-Lipschitz neural networks.
At the very core of our approach lies a new $1$-Lipschitz non-linear operator that we call {\em Convex Potential Layer} which allows us to adapt convex potential flows to the discretized case. 
These blocks enjoy the desirable property of stabilizing the training of the neural network by controlling the gradient norm, hence overcoming the exploding gradient issue.
We experimentally demonstrate our approach by training large-scale neural networks on several datasets, reaching state-of-the art results in terms of under-attack and certified accuracy.

\section{Background and Related Work}
\label{section:background_rw}

In this paper, we aim at devising {\em certified} defense mechanisms against adversarial attacks, in the following, we formally define an adversarial attacks and a robustness certificate.
We consider a classification task from an input space $\inputspace\subset\RR^d$ to a label space $\labelspace:=\{1,\dots,K\}$.
To this end, we aim at learning a classifier function $\mathbf{f}:=(f_1,\dots,f_K):\inputspace\to \RR^K$ such that the predicted label for an input $x$ is $\argmaxB_k f_k(x)$.
For a given couple input-label $(x,y)$, we say that $x$ is correctly classified if $\argmaxB_k f_k(x)=y$.

\begin{definition}[{\bf Adversarial Attacks}]
Let $x \in \mathcal{X}$, $y \in \mathcal{Y}$ the label of $x$ and let $\mathbf{f}$ be a classifier.
An adversarial attack at level $\varepsilon$ is a perturbation $\tau$ \st $\lVert\tau\rVert\leq\varepsilon$ such that:
\begin{equation*}
  \argmaxB_k f_k(x+\tau) \neq y
\end{equation*}
\end{definition}
Let us now define the notion of  robust certification. For $x \in \mathcal{X}$, $y \in \mathcal{Y}$ the label of $x$ and let $\mathbf{f}$ be a classifier, a classifier $\mathbf{f}$ is said to be \emph{certifiably robust at radius $\varepsilon\geq 0$} at point $x$ if for all $\tau$ such that ${\lVert\tau\rVert \leq \varepsilon}$ :
\begin{equation*}
  \argmaxB_k f_k(x+\tau) = y
\end{equation*}
The task of robust certification is then to find methods that ensure the previous property. A key quantity in this case is the Lipschitz constant of the classifier.

\subsection{Lipschitz property of Neural Networks}

The Lipschitz constant has seen a growing interest in the last few years in the field of deep learning~\citep{scaman2018lipschitz,fazlyab2019efficient,combettes2020lipschitz,bethune2021many}.
Indeed, numerous results have shown that neural networks with a small Lipschitz constant exhibit better generalization~\citep{bartlett2017spectrally}, higher robustness to adversarial attacks~\citep{szegedy2014intriguing,farnia2018generalizable,tsuzuku2018lipschitz}, better training stability~\citep{xiao2018dynamical,trockman2021orthogonalizing}, improved Generative Adversarial Networks~\citep{arjovsky2017wasserstein}, etc.
Formally, we define the Lipschitz constant with respect to the $\ell_2$ norm of a Lipschitz continuous function $f$ as follows:
\begin{equation*}
  \Lip_{2}{(f)} = \sup_{\substack{x, x' \in \mathcal{X} \\ x \neq x'}} \frac{\lVert f(x) - f(x') \rVert_2}{\lVert x - x' \rVert_2} \enspace.
\end{equation*}

Intuitively, if a classifier is Lipschitz, one can bound the impact of a given input variation on the output, hence obtaining guarantees on the adversarial robustness.
We can formally characterize the robustness of a neural network with respect to its Lipschitz constant with the following proposition:
\begin{prop}[\citet{tsuzuku2018lipschitz}] \label{proposition:tsuzuku}
Let $\mathbf{f}$ be an $L$-Lipschitz continuous classifier for the $\ell_2$ norm.
Let $\varepsilon > 0$, $x \in \mathcal{X}$ and $y \in \mathcal{Y}$ the label of $x$.
If at point $x$, the margin $\mathcal{M}_{\mathbf{f}}(x)$ satisfies:
\begin{equation*}
  \mathcal{M}_{\mathbf{f}}(x):=\max(0,f_y(x)-\max_{y'\neq y}f_{y'}(x)) > \sqrt{2} L \varepsilon
\end{equation*}
then we have for every $\tau$ such that $\lVert \tau \rVert_2 \leq \varepsilon$:
\begin{equation*}
  \argmaxB_{k}f_k(x + \tau) = y
\end{equation*}
\end{prop}
From Proposition~\ref{proposition:tsuzuku}, it is straightforward to compute a robustness certificate for a given point.
Consequently, in order to build robust neural networks the margin needs to be large and the Lipschitz constant small to get optimal guarantees on the robustness for neural networks.

\subsection{Certified Adversarial Robustness}

Mainly two kinds of methods have been developed to come up with certified adversarial robustness.
The first category relies on randomization and consists of convolving the input with a predefined probability distribution during both training and inference phases.
Several works that rely on the method have proposed empirical~\cite{cao2017mitigating,liu2018towards,pinot2019theoretical,pinot2020randomization} and certified defenses~\cite{lecuyer2018certified,li2019certified,cohen2019certified,salman2019provably,yang2020randomized}. These methods are model-agnostic, in the sense they do not depend on the architecture of the classifier, and provide ``high probability'' certificates.
However, this approach suffers from significant impossibility results: the maximum radius that can be certified for a given smoothing distribution vanishes as the dimension increases~\cite{yang2020randomized}.
Furthermore, in order to get non-vacuous provable guarantees, such approaches often require to query the network hundreds of times to infer the label of a single image.
This computational cost naturally limits the use of these methods in practice.

The second approach directly exploits the Lipschitzness property with the design of built-in $1$-Lipschitz layers. Contrarily to previous methods,  these approaches provide deterministic guarantees.
Following this line, one can either normalize the weight matrices by their largest singular values making the layer $1$-Lipschitz, \emph{e.g.}~\citep{yoshida2017spectral,miyato2018spectral,farnia2018generalizable,anil2019sorting} or project the weight matrices on the Stiefel manifold \citep{li2019preventing,trockman2021orthogonalizing,skew2021sahil}.
The work of \citet{li2019preventing}, \citet{trockman2021orthogonalizing} and \citet{skew2021sahil} (denoted BCOP, Cayley and SOC respectively) are considered the most relevant approach to our work.
Indeed, their approaches consist of projecting the weights matrices onto an orthogonal space in order to preserve gradient norms and enhance adversarial robustness by guaranteeing low Lipschitz constants. 
While both works have similar objectives, their execution is different.
The BCOP layer (Block Convolution Orthogonal Parameterization) uses an iterative algorithm proposed by \citet{bjorck1971iterative} to orthogonalize the linear transform performed by a convolution.
The SOC layer (Skew Orthogonal Convolutions) uses the property that if $A$ is a skew symmetric matrix then $Q=\exp{A}$ is an orthogonal matrix. To approximate the exponential, the authors proposed to use a finite number of terms in its Taylor series expansion.
Finally, the method proposed by~\citet{trockman2021orthogonalizing} use the Cayley transform to orthogonalize the weights matrices.
Given a skew symmetric matrix $A$, the Cayley transform consists in computing the orthogonal matrix $Q = (I - A)^{-1} (I + A)$. Both methods are well adapted to convolutional layers and are able to reach high accuracy levels on CIFAR datasets. Also, several works~\cite{anil2019sorting,singla2021householder,huang2021local} proposed methods leveraging the properties of activation functions to constraints the Lipschitz of Neural Networks. These works are usually useful to help  improving the performance of linear orthogonal layers.

\subsection{Residual Networks}

To prevent from gradient vanishing issues in neural networks during the training phase~\citep{hochreiter2001gradient},~\cite{he2016deep} proposed the Residual Network (ResNet) architecture.
Based on this architecture, several works~\citep{haber2017stable,e17Proposal,lu18beyond,chen2018neural} proposed a ``continuous time'' interpretation inspired by dynamical systems that can be defined as follows.

\begin{definition}\label{def:flow}
Let $(F_{t})_{t\in[0,T]}$ be a family of functions on $\RR^d$, we define the continuous time Residual Networks flow associated with $F_t$ as:
\begin{align*}\label{eq:resnet_c0}
  \left\{
    \begin{array}{ll}
    x_0 &= x\in\mathcal{X}\\
    \frac{dx_{t}}{dt} &= F_{{t}}(x_{t}) \  \text{for } \ t\in[0, T]
  \end{array}
  \right.
\end{align*}
\end{definition}

This continuous time interpretation helps as it allows us to consider the stability of the forward propagation through the stability of the associated dynamical system.
A dynamical system is said to be \emph{stable} if two trajectories starting from an input and another one remain sufficiently close to each other all along the propagation.
This stability property takes all its sense in the context of adversarial classification.

It was argued by~\citet{haber2017stable} that when $F_{t}$ does not depend on $t$ or vary slowly with time\footnote{This blurry definition of "vary slowly" makes the property difficult to apply.}, the stability can be characterized by the eigenvalues of the Jacobian matrix $\nabla_x F_{t}(x_t)$: 
the dynamical system is stable if the real part of the eigenvalues of the Jacobian stay negative throughout the propagation.
This property however only relies on intuition and this condition might be difficult to  verify in practice.
In the following, in order to derive stability properties, we study gradient flows and convex potentials, which are sub-classes of Residual networks.

Other works~\citep{huang2020adversarial,li2020implicit} also proposed to enhance adversarial robustness using dynamical systems interpretations of Residual Networks. Both works argues that using particular discretization scheme would make gradient attacks more difficult to compute due to numerical stability. These works did not provide any provable guarantees for such approaches.

\section{A Framework to design Lipschitz Layers}
\label{section:global_framework}

The continuous time interpretation of Definition~\ref{def:flow} allows us to better investigate the robustness properties and assess how a difference of the initial values (the inputs) impacts the inference flow of the model. Let us consider two continuous flows $x_t$ and $z_t$ associated with $F_t$ but differing in their respective initial values $x_0$ and $z_0$. Our goal is to characterize the time evolution of $\lVert x_t-z_t \rVert$ by studying its  time derivative. We recall that  every matrix $M\in\RR^{d\times d}$ can be uniquely decomposed as the sum of a symmetric and skew-symmetric matrix $M = S(M) + A(M)$. By applying this decomposition to the Jacobian matrix $\nabla_x F_t(x)$ of $F_t$, we can show that the time derivative of $\lVert x_t-z_t \rVert$ only involves the symmetric part  $S(\nabla_x F_t(x))$ (see Appendix~\ref{proof:continuous-lip} for details). 

For two symmetric matrices $S_1,S_2\in\RR^{d\times d}$,  we denote $S_1\preceq S_2$ if, for all $x\in\RR^d$, $\langle x,(S_2-S_1)x\rangle\geq 0$. By focusing on the symmetric part of the Jacobian matrix we can show in Appendix~\ref{proof:continuous-lip} the following proposition.
\begin{prop}
\label{prop:continuous-lip}
Let $(F_{t})_{t\in[0,T]}$ be a family of differentiable  functions almost everywhere on $\RR^d$.
Let us assume that there exists two measurable functions $t\mapsto \mu_t$ and  $t\mapsto \lambda_t$ such that
$$\mu_t I\preceq S(\nabla_xF_{t}(x))\preceq \lambda_tI$$
for all $x\in\RR^d$, and $t\in [ 0,T]$. Then the flow associated with $F_t$ satisfies for all initial conditions $x_0$ and $z_0$:
\begin{align*}
  \lVert x_0-z_0 \rVert e^{\int_0^t\mu_s ds}\leq \lVert x_t-z_t \rVert\leq \lVert x_0-z_0 \rVert e^{\int_0^t\lambda_s ds}
\end{align*}
\end{prop}

The symmetric part plays even a more important role since one can show that a function whose Jacobian is always skew-symmetric is actually linear (see Appendix~\ref{sup:skew} for more details). However, constraining $S(\nabla_x F_{t}(x))$ in the general case is technically difficult and a solution resorts to a more intuitive parametrization of  $F_t$ as the sum of two functions $F_{1,t}$ and $F_{2,t}$ whose Jacobian matrix are respectively symmetric  and skew-symmetric.  Thus, such a parametrization enforces $F_{2,t}$  to be linear and skew-symmetric. For the choice of $F_{1,t}$, we propose to rely on potential functions: a function  $F_{1,t}:\RR^d \to \RR^d$ derives from a simpler family of scalar valued function in $\RR^d$, called the \emph{potential}, via the gradient operation. Moreover, since the Hessian of the potential is symmetric, the Jacobian for $F_{1,t}$ is then also symmetric.  If we had the convex property to this potential, its Hessian has positive eigenvalues. Therefore we introduce the following corollary. See proof in Appendix~\ref{proof:conv-skew} 

\begin{corollary} 
\label{cor:conv-skew}Let $(f_{t})_{t\in[0,T]}$ be a family of convex differentiable functions on $\RR^d$ and $(A_t)_{t\in[0,T]}$ a family of skew symmetric matrices. Let us define 
$$F_t(x) = -\nabla_x f_{t}(x)+A_t x,$$ 
then the flow associated with $F_t$ satisfies for all initial conditions $x_0$ and $z_0$:
\begin{align*}
\lVert x_t-z_t \rVert\leq \lVert x_0-z_0 \rVert
\end{align*}
\end{corollary}

This simple property suggests that if we could parameterize $F_t$  with convex potentials, it would be less sensitive to input perturbations and therefore more robust to adversarial examples. We also remark that the skew symmetric part is then norm-preserving.
However, the discretization of such flow is challenging in order to maintain this property of stability.

\subsection{Discretized Flows}

To study the discretization of  the previous flow, let $t=1,\dots,T$ be the discretized time steps and from now we consider the flow defined by  $F_t(x) = -\nabla f_{t}(x)+A_t x$, with $(f_{t})_{t=1,\dots,T}$  a family of convex differentiable functions on $\RR^d$ and $(A_t)_{t=1,\dots,T}$ a family of skew symmetric matrices. The most basic method the explicit Euler scheme as defined by: 
\begin{align*}
x_{t+1} = x_t+ F_t(x_t)
\end{align*}
However, if $A_t\neq 0$, this discretized system might not satisfy $\lVert x_t-z_t\rVert\leq\lVert x_0-z_0\rVert$. Indeed, consider the simple example where $f_t=0$. We then have:
\begin{align*}
\lVert x_{t+1}-z_{t+1}\rVert - \lVert x_{t}-z_{t}\rVert =\lVert A_t\left(x_{t}-z_{t}\right)\rVert.
\end{align*}
Thus explicit Euler scheme cannot guarantee Lipschitzness when $A_t\neq 0$. To overcome this difficulty, the discretization step can be split in two parts, one for $\nabla_x f_t$ and one for $A_t$: 
\begin{align*}
   \left\{
    \begin{array}{ll}
        x_{t+\frac12} &= \textsc{step1}(x_t, \nabla_x f_t)\\
        x_{t+1}& = \textsc{step2}(x_{t+\frac12}, A_t)
    \end{array} 
    \right.
\end{align*}
This type of discretization scheme  can be found for instance from Proximal Gradient methods where one step is explicit and the other is implicit. Then, we dissociate the Lipschitzness study of both terms of the flow. 

\subsection{Discretization scheme for $\nabla_x f_t$}

To apply the explicit Euler scheme to $\nabla_x f_t$, an  additional smoothness property on the potential functions is required  to generalize the Lipschitzness guarantee to the discretized flows. Recall that a function $f$ is said to be \emph{$L$-smooth} if it is differentiable and if $x\mapsto\nabla_x f(x)$ is $L$-Lipschitz. 
\begin{prop}\label{prop:discrete_convex_potentials}
Let $t\in\{1,\cdots,T\}$ Let us assume that $f_{t}$ is $L_t$-smooth. We  define the following discretized ResNet gradient flow using $h_t$ as a step size:
\begin{align*}
    \begin{array}{ll}
    x_{t+\frac12} &= x_{t}-h_{t}\nabla_xf_{t}(x_{t})\\
  \end{array}
 \end{align*}
Consider now two trajectories $x_t$ and $z_t$ with initial points $x_0=x$ and $z_0=z$ respectively,  if $0\leq h_t\leq \frac{2}{L_t}$,  then 
$$\lVert x_{t+\frac12}-z_{t+\frac12}\rVert_2\leq \lVert x_t-z_t\rVert_2$$
\end{prop}
In Section~\ref{sec:param}, we describe how to parametrize a neural network layer to implement such a discretization step by leveraging the recent work on Input Convex Neural Networks~\cite{amos2017input}. 

\begin{rmq}
Another solution relies on the implicit Euler scheme: $ x_{t+\frac12} = x_{t}-\nabla_xf_{t}(x_{t+\frac12})$. We show in Appendix~\ref{sup:implicit} that this strategy defines a $1$-Lipschitz flow without further assumption on $f_t$ than convexity. We propose an implementation. However preliminary experiments did not show competitive results and the training time is prohibitive. We leave this solution for future work. 

\end{rmq}
\subsection{Discretization scheme for $A_t$}

The second step of discretization involves the term with skew-symmetric matrix $A_t$. As mentioned earlier, the challenge is that the \emph{explicit Euler discretization} is not contractive. More precisely,  the following property 
$$\lVert x_{t+1} - z_{t+1}\rVert\geq \lVert x_{t+\frac12} - z_{t+\frac12}\rVert$$ 
is satisfied with equality only in the special and useless case of $x_{t+\frac12} - z_{t+\frac12} \in \text{ker}(A_t)$. Moreover, the implicit Euler discretization induces an increasing norm and hence does not satisfy the desired property of norm preservation neither. 

\paragraph{Midpoint Euler method.}
We thus propose to use \emph{Midpoint Euler} method, defined as follows:
\begin{align*}
&x_{t+1} = x_{t+\frac12} +A_t \frac{x_{t+1}+x_{t+\frac12}}{2}\\
\iff&x_{t+1} = \left(I-\frac{A_t}{2}\right)^{-1}\left(I+\frac{A_t}{2}\right)x_{t+\frac12}.
\end{align*} 
Since $A_t$ is skew-symmetric, $I-\frac{A_t}{2}$ is invertible. This update corresponds to the Cayley Transform of $\frac{A_t}{2}$ that induces an orthogonal mapping. 
This kind of layers was introduced and extensively studied in~\citet{trockman2021orthogonalizing}.

\paragraph{Exact Flow.} One can define the simple differential equation corresponding to the flow associated with $A_t$  
\begin{align*}
        \frac{du_t}{ds} = A_t u_s,\quad u_0 = x_{t+\frac12},
\end{align*}
There exists an exact solution exists since $A_t$ is linear. By taking the value at $s=\frac12$, we obtained the following transformation:  
 \begin{align*}
x_{t+1} := u_{\frac12}=e^{\frac{A}{2}} x_{t+\frac12}.
\end{align*}
This step is therefore clearly norm preserving but the matrix exponentiation is challenging and it requires efficient approximations. This trend was recently investigated under the name of Skew Orthogonal Convolution (SOC)~\cite{skew2021sahil}.

\section{Parametrizing Convex Potentials Layers}
\label{sec:param}

As presented in the previous section, parametrizing the skew symmetric updates has been extensively studied by~\citet{trockman2021orthogonalizing,skew2021sahil}. In this paper we focus on  the parametrization of symmetric update with the convex potentials proposed in~\ref{prop:discrete_convex_potentials}. For that purpose, the Input Convex Neural Network (ICNN) \citep{amos2017input} provide a relevant starting point that we will extend. 

\subsection{Gradient of ICNN}
We use $1$-layer ICNN~\citep{amos2017input} to define an efficient computation of Convex Potentials Flows. For any vectors $w_1,\dots w_k\in\mathbb{R}^d$, and bias terms  $b_1,\dots,b_k\in \mathbb{R}$, and for $\phi$ a convex function,  the potential $F$ defined as:
\begin{align*}
    F_{w,b}:x\in\RR^d\mapsto \sum_{i=1}^k\phi( w_i^\top x+b_i)
\end{align*}
defines  a convex function in $x$ as the composition of a linear and a convex function. Its gradient with respect to its input $x$ is then:
\begin{align*}
    x\mapsto \sum_{i=1}^kw_i\phi'(w_i^\top x+b_i) = \mathbf{W}^\top \phi'(\mathbf{W} x+\mathbf{b})
\end{align*}
with $\mathbf{W}\in \mathbb{R}^{k\times d}$ and $\mathbf{b}\in\mathbb{R}^{k}$ are respectively the matrix and vector obtained by the concatenation of, respectively, $w_i^\top$ and $b_i$, and $\phi'$ is applied element-wise.  
Moreover, assuming $\phi'$ is $L$-Lipschitz, we have that $F_{w,b}$ is  $L\lVert\mathbf{W}\rVert_2^2$-smooth. $\lVert\mathbf{W}\rVert_2$ denotes the spectral norm of $\mathbf{W}$, \ie, the greatest singular value of $\mathbf{W}$ defined as:
\begin{align*}
   \lVert\mathbf{W}\rVert_2 :=\max_{x\neq 0} \frac{\lVert \mathbf{W}x\rVert_2}{\lVert x\rVert_2}
\end{align*}
The reciprocal also holds: if $\sigma:\RR\to\RR$ is a non-decreasing $L$-Lipschitz function, $\mathbf{W}\in \RR^{k\times d}$ and $b\in \RR^{k}$, there exists a convex $L\lVert\mathbf{W}\rVert_2^2$-smooth function $F_{w,b}$ such that 
$$\nabla_xF_{w,b}(x) =  \mathbf{W}^\top \sigma(\mathbf{W} x+\mathbf{b}),$$ where $\sigma$ is applied element-wise. The next section shows how this property can be used to implement the building block and training of such layers.

\subsection{Convex Potential layers}
From the previous section, we derive the following \emph{Convex Potential Layer}: 
\begin{equation*}
\label{equation:stable_block}
  z = x - \frac{2}{\lVert \mathbf{W} \lVert_2^2} \mathbf{W}^\top \sigma(\mathbf{W} x + b)
\end{equation*}
Written in a matrix form, this layer can be implemented with every linear operation $\mathbf{W}$.
In the context of image classification, it is beneficial to use convolutions\footnote{For instance, one can leverage the \texttt{Conv2D} and \texttt{Conv2D\_transpose} functions of the PyTorch framework~\citep{paszke2019pytorch}} instead of generic linear transforms represented by a dense matrix. 

\begin{rmq}
When $\mathbf{W}\in\RR^{1\times d}$, $b =0$ and $\sigma=\textsc{ReLU}$, the \emph{Convex Potential Layer} is equivalent to the HouseHolder activation function introduced in~\citet{singla2021householder}.
\end{rmq}

Residual Networks~\citep{he2016deep} are also composed of other types of layers which increase or decrease the dimensionality of the flow.
Typically, in a classical setting, the number of input channels is gradually increased, while the size of the image is reduced with pooling layers.
In order to build a $1$-Lipschitz Residual Network, all operations need to be properly scale or normalize in order to maintain the Lipschitz constant.

\paragraph{Increasing dimensionsionality.} To increase the number of channels in a convolutional Convex Potential Layer, a zero-padding operation can be easily performed: an input $x$ of size $c\times h \times w$ can be extended to some $x'$ of size  $c'\times h \times w$, where $c'>c$, which equals $x$ on the $c$ first channels and $0$ on the $c'-c$ other channels.
\paragraph{Reducing dimensionsionality.} Dimensionality reduction is another essential operation in neural networks. On one hand, its  goal is to  reduce the number of parameters and thus the amount of computation required to build the network. On the other hand it allows the model to progressively map the input space on the output dimension, which corresponds in many cases to the number of different labels $K$. 
In this context, several operations exist:
pooling layers are used to extract information present in a region of the feature map generated by a convolution layer. One can easily adapt pooling layers (\emph{e.g.} max and average) to make them $1$-Lipschitz~\citep{bartlett2017spectrally}.
Finally, a simple method to reduce the dimension is the product with a non-square matrix. In this paper, we simply implement it as  the truncation of the output. This obviously maintains the Lipschitz constant.

\begin{algorithm}[tb]
\caption{Computation of a Convex Potential Layer}
\label{algorithm:stable_block}
\begin{algorithmic}
  \STATE{Require: \bfseries Input: $x$, vector: $u$, weights: $\mathbf{W}$, $b$}
  \STATE{Ensure: Compute the layer $z$ and return $u$}
  \STATE{$v \gets \mathbf{W} u / \lVert \mathbf{W} u \rVert_2$}
  \STATE{$u \gets \mathbf{W}^\top v / \lVert \mathbf{W}^\top v \rVert_2$
    \rlap{\hspace{0.5cm}\smash{$\left.\begin{array}{@{}c@{}}\\{}\\{}\end{array}\right\}%
      \begin{tabular}{l}1 iter. for training \\100 iter. for inference\end{tabular}$}}}
  \STATE{$h \gets 2 / \left( \sum_i (\mathbf{W} u \cdot v)_i \right)^2$}
  \STATE{\textbf{return} $x - h \left[ \mathbf{W}^\top \sigma( \mathbf{W} x + b) \right], u$}
\end{algorithmic}
\end{algorithm}

\subsection{Computing spectral norms}
Our Convex Potential Layer, described in Equation~\ref{equation:stable_block}, can be adapted to any kind of linear transformations (\emph{i.e.} Dense or Convolutional) but requires the computation of the spectral norm for these transformations.
Given that computation of the spectral norm of a linear operator is known to be NP-hard~\citep{steinberg2005computation}, an efficient approximate method is required during training to keep the complexity tractable.

Many techniques exist to approximate the spectral norm (or the largest singular value), and most of them exhibit a trade-off between efficiency and accuracy.
Several methods exploit the structure of convolutional layers to build an upper bound on the spectral norm of the linear transform performed by the convolution~\citep{jia2017improving,singla2021fantastic,araujo2021lipschitz}.
While these methods are generally efficient, they can less relevant and adapted to certain settings. For instance in our context, using a loose upper bound of the spectral norm will hinder the expressive power of the layer and make it too contracting.

For these reasons we rely on the Power Iteration Method (PM).  This method converges at a geometric rate towards the largest singular value of a matrix. More precisely the convergence rate for a given matrix $\mathbf{W}$ is $\textstyle O((\frac{\lambda_2}{\lambda_1})^k)$ after $k$ iterations, independently from the choice for the starting vector, where $\lambda_1>\lambda_2$ are the two largest singular values of $\mathbf{W}$. While it can appear to be computationally expensive due to the large number of required iterations for convergence, it is possible to drastically reduce the number of iterations during training. Indeed, as in~\citep{miyato2018spectral}, by considering that the weights' matrices $\mathbf{W}$ change slowly during training, one can perform only one iteration of the PM for each step of the training and let the algorithm slowly converges along with the training process\footnote{Note that a typical training requires approximately 200K steps where 100 steps of PM is usually enough for convergence}.
We describe with more details in Algorithm~\ref{algorithm:stable_block}, the operations performed during a forward pass with a Convex Potential Layer. 

However for evaluation purpose, we need to compute the certified adversarial robustness, and this requires to ensure the convergence of the PM. Therefore, we perform $100$ iterations for each layer\footnote{$100$ iterations of Power Method is sufficient to converge with a geometric rate.} at inference time. Also note that at inference time, the computation of the spectral norm only needs to be performed once for each layer.

\section{Experiments}
\label{section:experiments}
\begin{table}
\begin{center}
    \begin{tabular}{cccccccc}
    \toprule
    \textbf{\#} & \textbf{S} &  & \textbf{M} & & \textbf{L} & & \textbf{XL} \\
    \midrule
    \textbf{Conv. Layers}      & 20 & & 30 & & 90 & & 120 \\
    \textbf{Channels}  &45 & & 60 & & 60 & & 70 \\ 
    \textbf{Lin. Layers}        &7 & & 10 & & 15 & & 15 \\
    \textbf{Lin. Features} & 2048 & & 2048 & & 4096 & & 4096 \\
    \bottomrule
    \end{tabular}%

\end{center}
\caption{\label{table:model-desc}
Architectures description for our Convex Potential Layers (CPL) neural networks with different capacities. We vary the number of Convolutional Convex Potential Layers, the number of Linear Convex Potential Layers, the number of channels in the convolutional layers and the width of fully
connected layers. In the paper, they will be reported respectively as CPL-S, CPL-M, CPL-L and CPL-XL.}
\end{table}

To evaluate our new $1$-Lipschitz Convex Potential Layers, we carry out an extensive set of experiments. In this section, we first describe  the details of our experimental setup. We then recall  the concurrent approaches that build $1$-Lipschitz Neural Networks and stress their limitations. Our experimental results are finally summarized in ection~\ref{sec:setting-xp}. By computing the certified and empirical adversarial  accuracy of our networks on CIFAR10 and CIFAR100 classification tasks~\citep{krizhevsky2009learning}, we show that our architecture is competitive with state-of-the-art methods (Sections~\ref{sec:results}). In Appendix~\ref{app:xp-supp}, we also study the influence of some hyperparameters and demonstrate the stability and the scalability of our approach by training very deep neural networks up to 1000 layers without normalization tricks or gradient clipping.

\subsection{Training and Architectural Details}
\label{sec:setting-xp}

We demonstrate the effectiveness of our approach on a classification task with CIFAR10 and CIFAR100 datasets~\citep{krizhevsky2009learning}. We use a similar training configuration to the one proposed in~\citep{trockman2021orthogonalizing}
We trained our networks with a batch size of $256$ over $200$ epochs.
We use standard data augmentation (\ie, random cropping and flipping), a learning rate of $0.001$ with Adam optimizer \citep{diederik2014adam} without weight decay and a piecewise triangular learning rate scheduler. We used a margin parameter in the loss set to $0.7$.

As other usual convolutional neural networks, we first stack few Convolutional CPLs and then stack some Linear CPLs for classification tasks. To validate the performance  and the scalability of our layers,  we will evaluate four different variations of different hyperparameters as described in Table~\ref{table:model-desc}, respectively named CPL-S, CPL-M, CPL-L and CPL-XL, ranked according to the number of parameters they have. In all our experiments, we made $3$ independent trainings to evaluate accurately the models. All reported results are the average of these $3$ runs.

\subsection{Concurrent Approaches} We compare our networks with SOC~\citep{skew2021sahil} and Cayley~\cite{trockman2021orthogonalizing} networks which are to our knowledge the best performing approaches for deterministic $1$-Lipschitz Neural Networks. Since our layers are fundamentally different from these ones, we cannot compare with the same architectures. We reproduced SOC results for with $10$ and $20$ layers, that we call respectively SOC-$10$ and SOC-$20$ in the same training setting, \emph{i.e.} normalized inputs, cross entropy loss, SGD optimizer with learning rate $0.1$ and multi-step learning rate scheduler. For Cayley layers networks, we reproduced their best reported model, \emph{i.e.} KWLarge with width factor of $3$. 

The work of~\citet{singla2021householder} propose three methods to improve certifiable accuracies from SOC layers: a new HouseHolder activation function (HH),  last layer normalization (LLN), and certificate regularization (CR). The code associated with this paper is not open-sourced yet, so we just reported the results from their paper in ours results (Tables~\ref{table:c10-comp} and~\ref{table:c100-comp}) under the name SOC+. We were being able to implement the LLN method in all models. This method largely improve the result of all methods on CIFAR100, so we used it for all networks we compared on CIFAR100 (ours and concurrent approaches).

\begin{table*}[tb]
  \centering
  \sisetup{%
    table-align-uncertainty=true,
    separate-uncertainty=true,
    detect-weight=true,
    detect-inline-weight=math
  }
  \begin{tabular}
  {
    l
    S[table-format=2.2]
    S[table-format=2.2]
    S[table-format=2.2]
    S[table-format=2.2]
    S[table-format=2.2]
  }
  \toprule
    & \multicolumn{1}{c}{\textbf{Clean Accuracy}} & \multicolumn{3}{c}{\textbf{Provable Accuracy ($\varepsilon $)}} &  \multicolumn{1}{c}{\textbf{Time per epoch (s)}} 
    \\
    \cmidrule{3-5}
    & \multicolumn{1}{c}{ } & \multicolumn{1}{c}{36/255} & \multicolumn{1}{c}{72/255} &  \multicolumn{1}{c}{108/255} & \multicolumn{1}{c}{\textbf{}} 
    \\
  \midrule
    \textbf{CPL-S} & 75.6  & 62.3  & 46.9  & 32.2  & 21.9 \\
  \textbf{CPL-M} & 76.8  & 63.3  & 47.5  & 32.5  & 40.0 \\
  \textbf{CPL-L} & 77.7  & 63.9 & 48.1 & 32.9  & 93.4 \\
  \textbf{CPL-XL} & 78.5  & 64.4  & 48.0  & 33.0 & 163 \\
  \midrule
  \textbf{Cayley (KW3)} & 74.6  & 61.4  & 46.4  & 32.1  & 30.8\\
    \midrule

  \textbf{SOC-10} & 77.6  & 62.0  & 45.0  & 29.5  & 33.4 \\
  \textbf{SOC-20} & 78.0  & 62.7 & 46.0  & 30.3  &52.2\\
  \midrule
    \textbf{SOC+-10} & 76.2 &62.6 & 47.7 & 34.2& N/A\\
  \textbf{SOC+-20} & 76.3&62.6& 48.7& 36.0& N/A \\

  \bottomrule
  \end{tabular}%
  \caption{Results on the CIFAR10 dataset on standard and  provably certifiable accuracies for different values of perturbations $\varepsilon$ on CPL (ours), SOC and Cayley models. The average time per epoch in seconds is also reported in the last column. None of these networks uses Last Layer Normalization.}
  \label{table:c10-comp}%
\end{table*}%

\begin{table*}[tb]
  \centering
  \sisetup{%
    table-align-uncertainty=true,
    separate-uncertainty=true,
    detect-weight=true,
    detect-inline-weight=math
  }
  \begin{tabular}
  {
    l
    S[table-format=2.2]
    S[table-format=2.2]
    S[table-format=2.2]
    S[table-format=2.2]
    S[table-format=2.2]
  }
  \toprule
    & \multicolumn{1}{c}{\textbf{Clean Accuracy}} & \multicolumn{3}{c}{\textbf{Provable Accuracy ($\varepsilon $)}} &  \multicolumn{1}{c}{\textbf{Time per epoch (s)}} 
    \\
    \cmidrule{3-5}
    & \multicolumn{1}{c}{ } & \multicolumn{1}{c}{36/255} & \multicolumn{1}{c}{72/255} &  \multicolumn{1}{c}{108/255} & \multicolumn{1}{c}{\textbf{}} 
    \\
  \midrule
    \textbf{CPL-S} & 44.0  & 29.9  & 19.1  & 11.0  & 22.4 \\
  \textbf{CPL-M} & 45.6  & 31.1  & 19.3  & 11.3 & 40.7 \\
  \textbf{CPL-L} & 46.7  & 31.8 & 20.1  & 11.7  & 93.8 \\
  \textbf{CPL-XL} & 47.8  & 33.4  & 20.9  &  12.6  & 164 \\
  \midrule
  \textbf{Cayley (KW3)} & 43.3  & 29.2  & 18.8 & 11.0  & 31.3 \\
    \midrule

  \textbf{SOC-10} & 48.2  & 34.3  &22.7 & 14.0  & 33.8 \\
  \textbf{SOC-20} & 48.3  & 34.4 & 22.7  & 14.2 & 52.7 \\
  \midrule
    \textbf{SOC+-10} & 47.1& 34.5& 23.5& 15.7& N/A \\
  \textbf{SOC+-20} & 47.8 & 34.8 & 23.7 & 15.8 &  N/A\\

  \bottomrule
  \end{tabular}%
  \caption{Results on the CIFAR100 dataset on standard and  provably certifiable accuracies for different values of perturbations $\varepsilon$ on CPL (ours), SOC and Cayley models. The average time per epoch in seconds is also reported in the last column. All the reported networks use Last Layer Normalization.}
  \label{table:c100-comp}%
\end{table*}%

\subsection{Results}
\label{sec:results}

In this section, we present our results on adversarial robustness.
We provide results on provable $\ell_2$ robustness as well as empirical robustness on CIFAR10 and CIFAR100 datasets for all our models and the concurrent ones

\begin{figure}[h]
    \centering
    \begin{tabular}{cc}
    \includegraphics[width =0.48\textwidth]{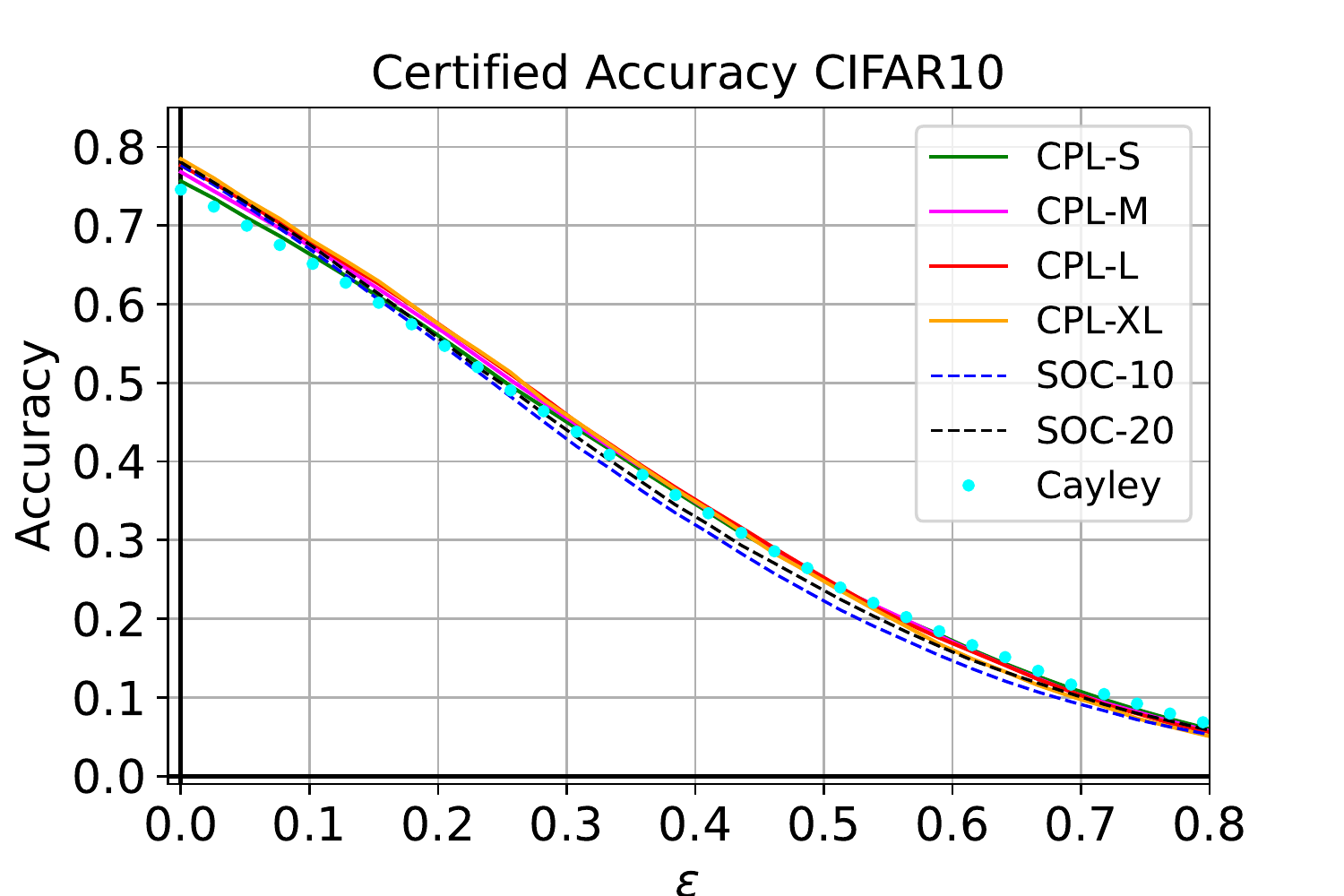} & \includegraphics[width =0.48\textwidth]{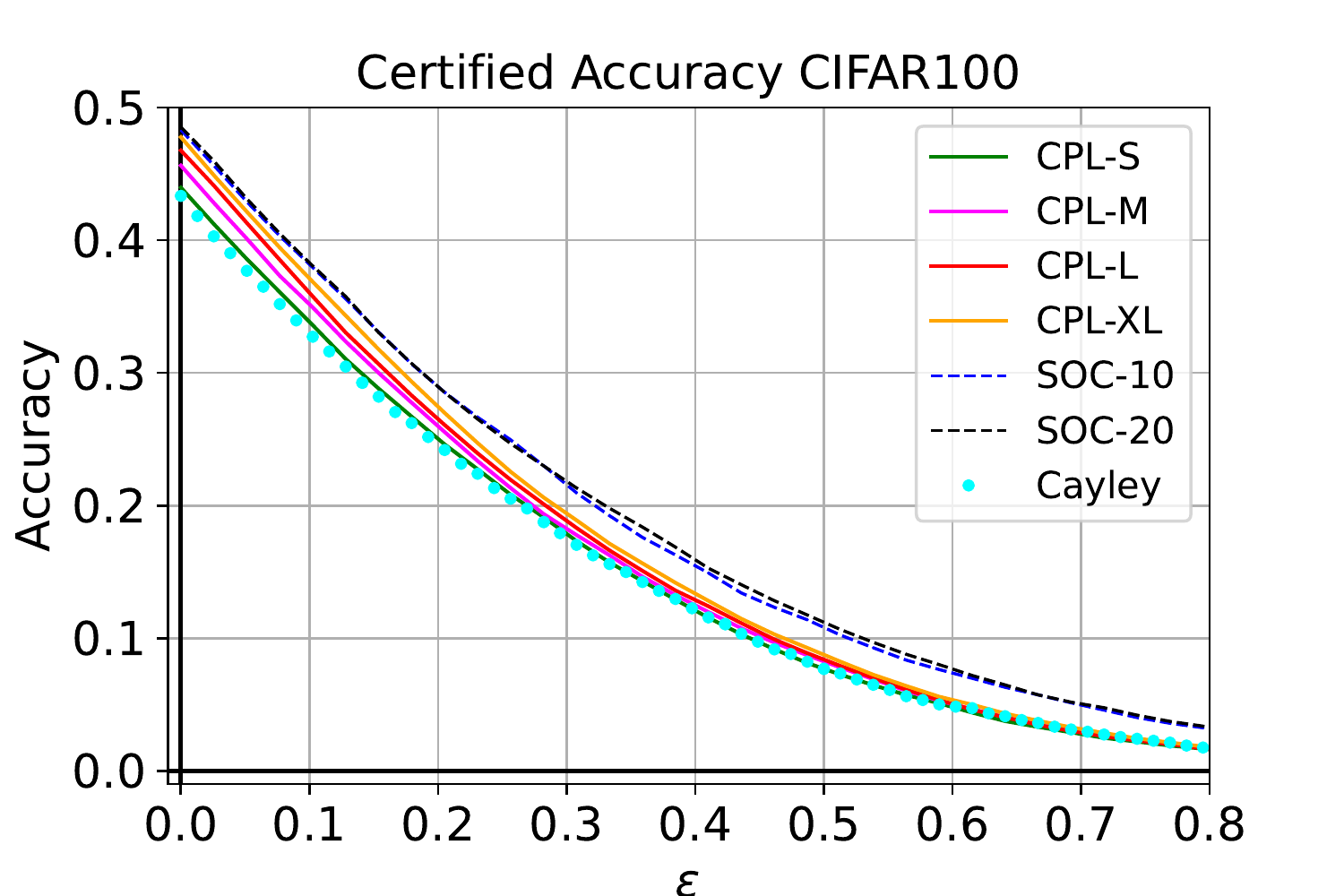}
    \end{tabular}
    \caption{Certifiably robust accuracy in function of the perturbation $\varepsilon$ for our CPL networks and its concurrent approaches (SOC and Cayley models) on CIFAR10 and CIFAR100 datasets.}
    \label{fig:cert-acc}
\end{figure}

\paragraph{Certified Adversarial Robustness.} 
Results on CIFAR10 and CIFAR100 dataset are reported respectively in Tables~\ref{table:c10-comp} and~\ref{table:c100-comp}. We also plotted certified accuracy in function of $\varepsilon$ on Figure~\ref{fig:cert-acc}. On CIFAR10, our method outperforms the concurrent approaches in terms of standard and certified accuracies for every level of $\varepsilon$ except SOC+ that uses additional tricks we did not use. On CIFAR100, our method performs slightly under the SOC networks but better than Cayley networks. Overall, our methods reach competitive results with SOC and Cayley layers. 

Note that we observe a small gain using larger and deeper architectures for our models. This gain is less important as $\varepsilon$ increases but the gain is non negligible for standard accuracies. In term of training time, our small architecture (CPL-S) trains very fast compared to other methods, while larger ones are longer to train.

\begin{figure}[h]
    \centering
    \begin{tabular}{cc}
    \includegraphics[width =0.48\textwidth]{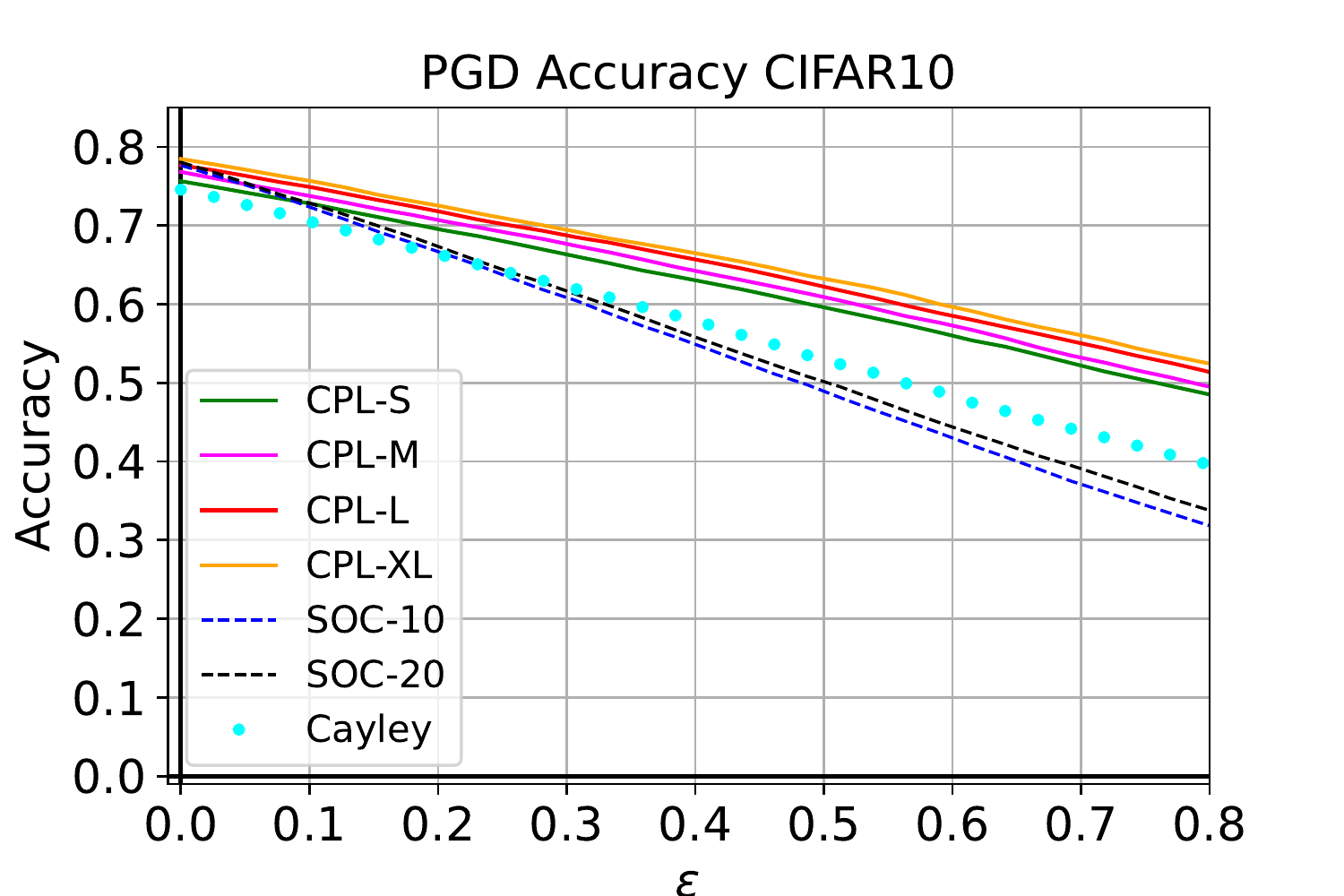} & \includegraphics[width =0.48\textwidth]{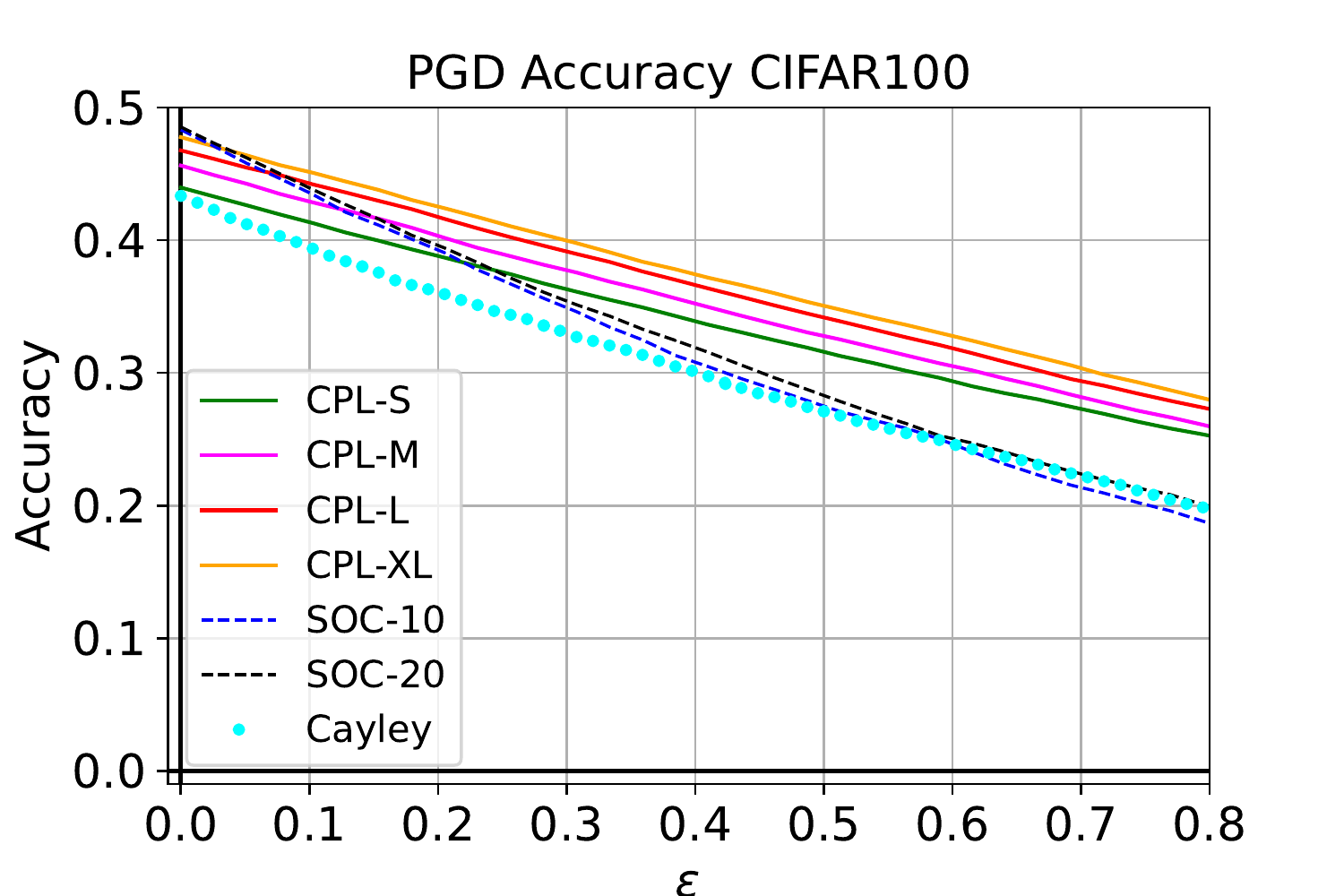}
    \end{tabular}
    \caption{Accuracy against PGD attack with 10 iterations in function of the perturbation $\varepsilon$ for our CPL networks and its concurrent approaches on CIFAR10 and CIFAR100 datasets.}
    \label{fig:pgd-acc}
\end{figure}
\paragraph{Empirical Adversarial Robustness.} We also reported in Figure~\ref{fig:pgd-acc} the accuracy of all the models against PGD $\ell_2$-attack~\citep{kurakin2016adversarial,madry2017towards} for various levels of $\epsilon$. We used $10$ iterations for this attack. We remark here that our methods brings a large gain of robust accuracy over all other methods. On CIFAR10 for $\varepsilon = 0.8$, the gain of CPL-S over SOC-10 approach is more than $10\%$. For CIFAR100, the gain is about $10\%$ too for $\varepsilon=0.6$. We remark that using larger architectures lead in a more substantial gain in empirical robustness. 

Our layers  only provide an upper bound on the Lipschitz constant, while orthonormal layers as Cayley and SOC are built to exactly preserve the norms. This might negatively influence the certified accuracy since the effective Lipschitz constant is smaller than the theoretical one, hence leading to suboptimal certificates. This might explain why our method performs so well of empirical robustness task.

\section{Conclusion}
In this paper, we presented a new generic method to build $1$-Lipschitz layers.
We leverage the continuous time dynamical system interpretation of Residual Networks and show that using convex potential flows naturally defines $1$-Lipschitz neural networks.
After proposing a parametrization based on Input Convex Neural Networks~\citep{amos2017input}, we  show that our models  reach competitive results in classification and robustness in comparison which other existing $1$-Lipschitz approaches.
We also experimentally show that our layers provide scalable approaches without further regularization tricks to train very deep architectures.

Exploiting the ResNet architecture for devising flows have been an important research topic.
For example, in the context of generative modeling, Invertible Neural Networks~\citep{behrmann2019invertible} and Normalizing Flows~\citep{rezende2015variational, verine2021expressivity} are both import research topic.
More recently, Sylvester Normalizing Flows~\citep{vdberg2018sylvester} or Convex Potential Flows~\citep{huang2021convex} have had similar ideas to this present work but for a very different setting and applications. In particular, they did not have interest in the contraction property of convex flows and the link with adversarial robustness have been under-exploited.

\paragraph{Further work.}
Propoisition~\ref{prop:continuous-lip} suggests to constraint the symmetric part of the Jacobian of $F_t$. We proposed to decompose $F_t$ as a sum of potential gradient and skew symmetric matrix. Finding other parametrizations is an open challenge.
Our models may not express all $1$-Lipschitz functions.
Knowing which functions can be approximated by our CPL layers is difficult even in the linear case (see Appendix~\ref{app:express}). This is  an important question that requires further investigation. 
One can also think of extending  our work by the study of  other dynamical systems. Recent architectures such as Hamiltonian Networks~\citep{greydanus2019hamiltonian} and Momentum Networks~\citep{sander2021momentum} exhibit interesting properties.
Finally, we hope to use similar approaches to build robust Recurrent Neural Networks~\citep{sherstinsky2020fundamentals} and Transformers~\citep{vaswani2017attention}.

\newpage
\bibliography{bibliography}
\bibliographystyle{plainnat}

\clearpage
\appendix
\onecolumn

\section{Further Related Work}

\paragraph{Lipschitz Regularization for Robustness.}

Based on the insight that Lipschitz Neural Networks are more robust to adversarial attacks, researchers have developed several techniques to regularize and constrain the Lipschitz constant of neural networks.
However the computation of the Lipschitz constant of neural networks has been shown to be NP-hard~\citep{scaman2018lipschitz}.  Most methods therefore tackle the problem by reducing or constraining the Lipschitz constant at the layer level.
For instance, the work of~\citet{cisse2017parseval,huang2020controllable} and ~\citet{wang2020orthogonal} exploit the orthogonality of the weights matrices to build Lipschitz layers.
Other approaches~\citep{gouk2021regularisation,jia2017improving,sedghi2018singular,singla2021fantastic,araujo2021lipschitz} proposed to estimate or upper-bound the spectral norm of convolutional and dense layers using for instance the power iteration method~\citep{golub2000eigenvalue}.
While these methods have shown interesting results in terms of accuracy, empirical robustness and efficiency, they can not provide provable guarantees since the Lipschitz constant of the trained networks remains unknown or vacuous.

\paragraph{Reshaped Kernel Methods.}
It has been shown by~\citet{cisse2017parseval} and \citet{tsuzuku2018lipschitz} that the spectral norm of a convolution can be upper-bounded by the norm of a reshaped kernel matrix. Consequently, orthogonalizing directly this matrix upper-bound the spectral norm of the convolution by $1$.
While this method is more computationally efficient than orthogonalizing the whole convolution, it lacks expressivity as the other singular values of the convolution are certainly too constrained.

\section{Proofs}

\subsection{Proof of Proposition~\ref{prop:continuous-lip}}
\label{proof:continuous-lip}

\begin{proof}
Consider the time derivative of the square difference between the two flows $x_t$ and $z_t$ associated with the function $F_t$ and following the definition~\ref{def:flow}: 
\begin{align*}
  \frac{d}{dt} \lVert x_t-z_t\rVert_2^2 & = 2 \big\langle x_t-z_t,\frac{d}{dt}( x_t-z_t)\big\rangle\\
    &=2 \big\langle x_t-z_t,F_{\theta_{t}}(x_{t})-F_{\theta_{t}}(z_{t})\big\rangle \\
    &=  2 \big\langle x_t-z_t,\int_0^1\nabla_xF_{\theta_{t}}(x_{t}+s(z_t-z_t))(x_t-z_t)ds\big\rangle\textrm{, by Taylor-Lagrange formula}\\ 
    &=  2 \int_0^1\big\langle x_t-z_t,\nabla_xF_{\theta_{t}}(x_{t}+s(z_t-z_t))(x_t-z_t)\big\rangle ds\\
     &=  2 \int_0^1\big\langle x_t-z_t,S(\nabla_xF_{\theta_{t}}(x_{t}+s(z_t-z_t)))(x_t-z_t)\big\rangle ds
\end{align*}
In the last step, we used that for every skew-symmetric matrix $A$ and vector $x$, $\lVert x,Ax\rVert = 0$.
Since $\mu_tI\preceq S(\nabla_xF_{\theta_{t}}(x_{t}+s(z_t-y_t)))\preceq  \lambda_tI$, we get
\begin{align*}
 2\mu_t \lVert x_t-z_t\rVert_2^2 \leq \frac{d}{dt} \lVert x_t-z_t\rVert_2^2 \leq 2\lambda_t \lVert x_t-z_t\rVert_2^2
\end{align*}
Then by Gronwall Lemma, we have
\begin{align*}
  \lVert x_0-y_0 \rVert e^{\int_0^t\mu_s ds}\leq \lVert x_t-y_t \rVert\leq \lVert x_0-y_0 \rVert e^{ \int_0^t\lambda_s ds}
\end{align*}
which concludes the proof.
\end{proof}

\subsection{Proof of Corollary~\ref{cor:conv-skew}}
\label{proof:conv-skew}
\begin{proof}

For all $t,x$, we have $F_t(x) = -\nabla_x f_{t}(x)+A_t x$ ~ so~
$\nabla_x F_t(x) = -\nabla_x^2 f_{t}(x)+A_t$. Then $S(\nabla_x F_t(x)) =-\nabla_x^2 f_{t}(x)$. Since $f$ is convex, we have $\nabla_x^2 f_{t}(x)\succeq 0$. So by application of Proposition~\ref{prop:continuous-lip}, we deduce $\lVert x_t-y_t \rVert\leq \lVert x_0-y_0 \rVert$ for all trajectories starting from $x_0$ and $y_0$.
\end{proof}

\subsection{Proof of Proposition~\ref{prop:discrete_convex_potentials}}
\label{proof:discrete_convex_potentials}
\begin{proof}
With $c_t = \lVert x_t -z_t\rVert_2^2$, we can write:
\begin{align*}
   c_{t+\frac12} - c_t = &-2 h_t \big\langle x_t - z_t, \nabla_xF_{\theta_{t}}(x_t) - \nabla_xF_{\theta_{t}}(z_t)  \big\rangle+ h_t^2 \lVert \nabla_xF_{\theta_{t}}(z_t) - \nabla_xF_{\theta_{t}}(z_t)\rVert_2^2
\end{align*}
This equality allows us to derive the equivalence between  $c_{t+1} \leq c_t$ and: 
\begin{align*}
   \frac{h_t}{2}
   \lVert  \nabla F_{\theta_{t}}(x_t) - \nabla F_{\theta_{t}}(z_t)\rVert_2^2
   \leq
   \langle x_t -z_t, \nabla F_{\theta_{t}}(z_t) - \nabla F_{\theta_{t}}(z_t) \rangle 
\end{align*}
Moreover, assuming that $F_{\theta_t}$ being  that:
\begin{align*}
   \frac{1}{L_t} &\lVert \nabla_xF_{\theta_{t}}(x_t) - \nabla_xF_{\theta_{t}}(z_t)\rVert_2^2 
   \leq\big\langle x_t -z_t, \nabla_xF_{\theta_{t}}(x_t) - \nabla_xF_{\theta_{t}}(z_t) \big\rangle
\end{align*}
We can see with this last inequality that if we enforce  $h_t \leq \frac{2}{L_t}$, we get $c_{t+\frac12} \leq c_t$ which concludes the proof.
\end{proof}

\section{Additional Results}

\subsection{Functions whose gradient is skew-symmetric everywhere}
\label{sup:skew}
Let $F:=(F_1,\dots,F_d):\RR^d\to\RR^d$ be a twice differentiable function such that $\nabla F(x)$ is skew-symmetric for all $x\in\RR^d$. Then we have for all $i,j,k$:
\begin{align*}
    \partial_i\partial_j F_k =  -\partial_i\partial_k F_j =-\partial_k\partial_i F_j = \partial_k\partial_j F_i = \partial_j\partial_k F_i = -\partial_j\partial_i F_k = -\partial_i\partial_j F_k
\end{align*}
So we have $\partial_i\partial_j F_k =0$ and then $F$ is linear: there exists a skew-symmetric matrix $A$ such that $F(x)=Ax$

\subsection{Implicit discrete convex potential flows}
\label{sup:implicit}

Let us define the implicit update $x_{t+\frac12} = x_{t}-\nabla_xf_{t}(x_{t+\frac12})$. Let us remark that $x_{t+\frac12}$ is uniquely defined as:
\begin{align*}
 x_{t+\frac12} =\argminB_{x\in\RR^d} \frac 12\lVert x-x_t\rVert^2+ f_t(x)  
\end{align*}
We recognized here the proximal operator of $f_t$ that is uniquely defined since $f_t$ is convex. Moreover we have for two trajectories $x_t$ and $z_t$:
\begin{align*}
   \lVert x_t-z_t\rVert^2_2& =  \lVert x_{t+\frac12}-z_{t+\frac12} +\nabla_xf_{t}(x_{t+\frac12})-\nabla_xf_{t}(z_{t+\frac12}) \rVert^2_2 \\
   &=  \lVert x_{t+\frac12}-z_{t+\frac12}\rVert^2 + 2\langle x_t-z_t,\nabla_xf_{t}(x_{t+\frac12})-\nabla_xf_{t}(z_{t+\frac12})\rangle +\lVert\nabla_xf_{t}(x_{t+\frac12})-\nabla_xf_{t}(z_{t+\frac12}) \rVert^2_2 \\
   &\geq  \lVert x_{t+\frac12}-z_{t+\frac12}\rVert^2 
\end{align*}
where the last inequality is deduced from the convexity of $f_t$. So, without any further assumption on $f_t$, the discretized implicit convex potential flow is $1$-Lipschitz.

To compute such a layer, one could solve the proximal operator strongly convex-minimization optimization problem. This strategy is not computationally efficient and not scalable.

\subsection{Expressivity of discretized convex potential flows}
\label{app:express}
Let us define $\mathcal{S}_1(\RR^{d\times d})$ the space of real symmetric matrices with singular values bounded by $1$. Let us also define $\mathcal{M}_1(\RR^{d\times d})$ the space of real matrices with singular values bounded by $1$ in absolute value.
Let $\mathcal{P}(\RR^{d\times d})=\{A\in\RR^{d\times d}|\exists n\in \mathbb{N},S_1,\dots,S_n\in \mathcal{S}_1(\RR^d\times d)\text{ s.t. } A = S_1\dots S_n\}$. Then one can prove\footnote{A proof and justification of this result can be found here: \url{https://mathoverflow.net/questions/60174/factorization-of-a-real-matrix-into-hermitian-x-hermitian-is-it-stable}} that $\mathcal{P}(\RR^{d\times d}) \neq \mathcal{M}_1(\RR^{d\times d})$. Thus there exists $A\in\mathcal{M}_1(\RR^{d\times d})$ such that for all matrices $n$, for all matrices $S_1,\dots,S_n\in\mathcal{S}_1(\RR^{d\times d})$ such that $M\neq S_1,\dots,S_n$. 

Applied to the expressivity of discretized convex potential flows, the previous result means that there exists a $1$-Lipschitz linear function that cannot be approximated as a discretized flow of any depth of convex linear $1$-smooth potential flows as in Proposition~\ref{prop:discrete_convex_potentials}. Indeed such a flow would write: $x\mapsto\prod_i(1-2S_i)x$ where $S_i$ are symmetric matrices whose eigenvalues are in $[0,1]$, in other words such transformations are exactly described by $x\mapsto Mx$  for some $M\in\mathcal{P}(\RR^{d\times d})$.

\section{Additional experiments}
\label{app:xp-supp}

\subsection{Training stability: scaling up to $1000$ layers}

\begin{figure}[h]
    \centering
    \includegraphics[width=0.5\textwidth]{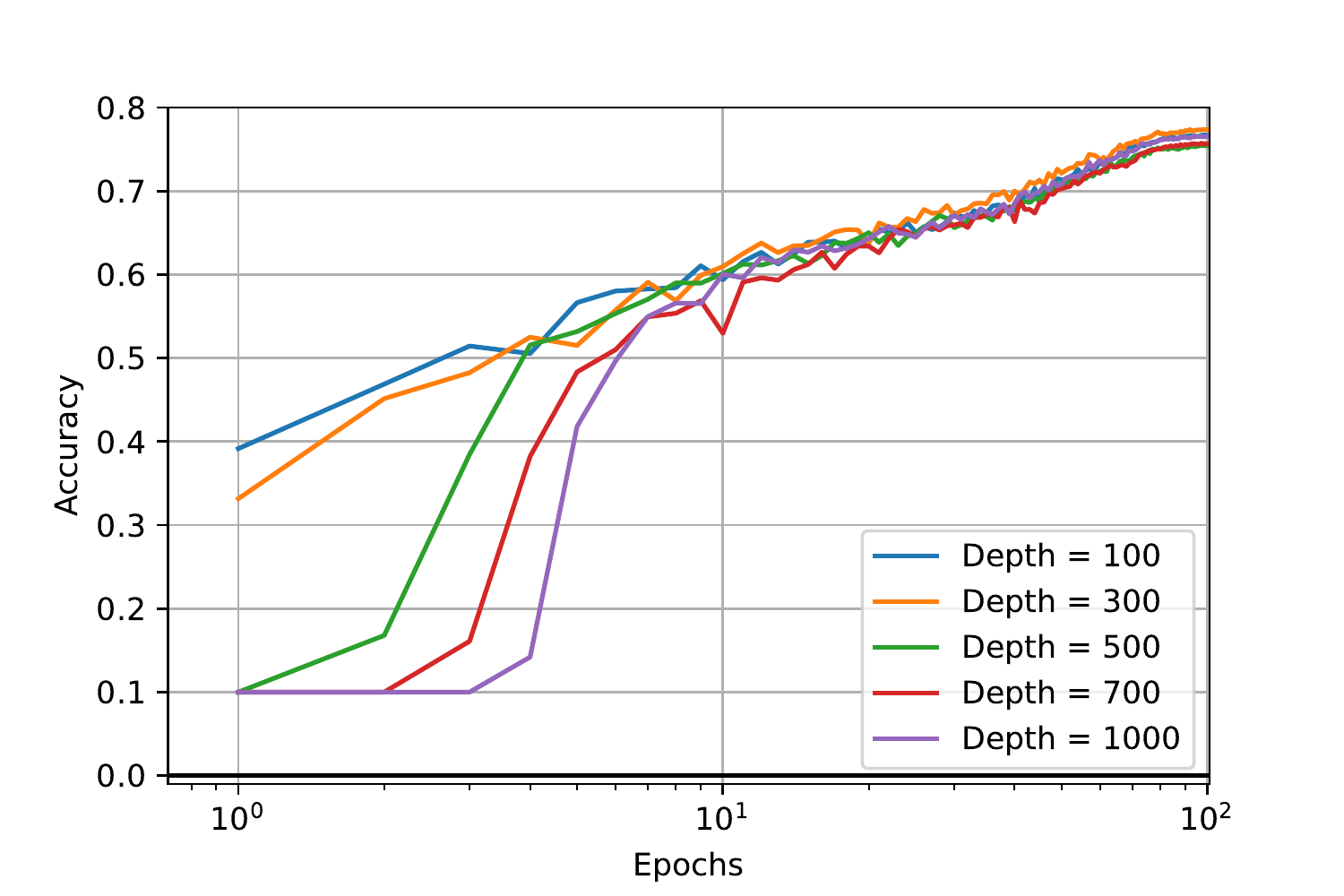}
    \caption{Standard test accuracy in function of the number of epochs (log-scale) for various depths for our neural networks ($100,300,500,700,1000$).}
    \label{fig:verydeep}
\end{figure}

While the Residual Network architecture limits, by design, gradient vanishing issues, it still suffers from exploding gradients in many cases~\citep{hayou2021stable}.
To prevent such scenarii, batch normalization layers~\citep{ioffe2015batch} are used in most Residual Networks to stabilize the training.

Recently, several works~\citep{miyato2018spectral,farnia2018generalizable} have proposed to normalize the linear transformation of each layer by their spectral norm.
Such a method would limit exploding gradients but would again suffer from gradient vanishing issues.
Indeed, spectral normalization might be too restrictive: dividing by the spectral norm can make other singular values vanishingly
small.
While more computationally expensive (spectral normalization can be done with $1$ Power Method iteration), orthogonal projections prevent both exploding and vanishing issues. 

On the contrary the architecture proposed in this paper has the advantage to naturally control the gradient norm of the output with respect to a given layer.
Therefore, our architecture can get the best of both worlds: limiting exploding and vanishing issues while maintaining scalability. 
To demonstrate the scalability of our approach, we experiment the ability to scale our architecture to very high depth (up to 1000 layers) without any additional normalization/regularization tricks, such as Dropout~\citep{srivastava2014dropout}, Batch Normalization~\citep{ioffe2015batch} or gradient clipping~\citep{pascanu2013difficulty}.
With the work done by~\cite{xiao2018dynamical}, which leverage Dynamical Isometry and a Mean Field Theory to train a $10000$ layers neural network, we believe, to the best of our knowledge, to be the second to perform such training. 
For sake of computation efficiency, we limit this experiment to architecture with $30$ feature maps.
We report the accuracy in terms of epochs for our architecture in Figure~\ref{fig:verydeep} for a varying number of convolutional layers.
It is worth noting that for the deepest networks, it may take a few epochs before the start of convergence.
As \cite{xiao2018dynamical}, we remark there is no gain in using very deep architecture for this task.

\subsection{Relaxing linear layers}

\begin{center}
\begin{tabular}{lrrr}
\toprule
  & \multicolumn{1}{c}{\textbf{h = 1.0}} & \multicolumn{1}{c}{\textbf{h = 0.1}} & \multicolumn{1}{c}{\textbf{h = 0.01}} \\
\midrule
\textbf{Clean} & 85.10 & 82.23 & 78.53 \\
\textbf{PGD ($\varepsilon = 36/255$}) & 61.45 & 62.99 & 60.98 \\
\bottomrule
\end{tabular}%
\end{center}
The table above shows the result of the relaxed training of our StableBlock architecture, i.e. we fixed the step $h_t$ in the discretized convex potential flow of Proposition~\ref{prop:discrete_convex_potentials}.
Increasing the constant $h$ allows for an important improvement  in the clean accuracy, but we loose in robust empirical accuracy.
While computing the certified accuracy is not possible in this case due to the unknown value of the Lipschitz constant, we can still notice that the training of the network are still stable without normalization tricks, and offer a non-negligible level of robustness.

\subsection{Effect of Batch Size in Training}

\begin{table*}[h]
  \centering
  \sisetup{%
    table-align-uncertainty=true,
    separate-uncertainty=true,
    detect-weight=true,
    detect-inline-weight=math
  }
  \begin{tabular}
  {
    l
    S[table-format=2.2]
    S[table-format=2.2]
    S[table-format=2.2]
    S[table-format=2.2]
    S[table-format=2.2]
    S[table-format=2.2]
  }
  \toprule
   &\multicolumn{1}{c}{\textbf{Batch size }}& \multicolumn{1}{c}{\textbf{Clean Accuracy}} & \multicolumn{3}{c}{\textbf{Provable Accuracy ($\varepsilon $)}} &  \multicolumn{1}{c}{\textbf{Time per epoch (s)}} 
    \\
    \cmidrule{4-6}
    & \multicolumn{1}{c}{ } & \multicolumn{1}{c}{ } &\multicolumn{1}{c}{36/255} & \multicolumn{1}{c}{72/255} &  \multicolumn{1}{c}{108/255} & \multicolumn{1}{c}{\textbf{}} 
    \\
  \midrule
    \multirow{3}{*}{\textbf{CPL-S}} & 64 & 76.5& 62.9 & 47.3 & 32.0 & 48 \\
                                    & 128 & 76.1 & 62.8 & 47.1 & 32.3  & 31 \\
                                    & 256 & 75.6 & 62.3 & 46.9 & 32.2 & 22 \\
    \midrule
 \multirow{3}{*}{\textbf{CPL-M}} & 64 & 77.4 & 63.6 & 47.4 & 32.1  & 77 \\
                                    & 128 & 77.2 &63.5 & 47.5 & 32.1 & 50 \\
                                    & 256 & 76.8 & 63.2 & 47.4 & 32.4& 40 \\
    \midrule

 \multirow{3}{*}{\textbf{CPL-L}} & 64 & 78.4 & 64.2 & 47.8 & 32.2  & 162 \\
                                    & 128 & 78.2 & 64.3 & 47.9 & 32.5 & 109 \\
                                    & 256 & 77.6 & 63.9 & 48.1 & 32.7& 93 \\
   \midrule
 \multirow{3}{*}{\textbf{CPL-XL}} & 64 & 78.9 & 64.2 & 47.2 & 31.2  & 271 \\
                                    & 128 & 78.9 & 64.2 & 47.5 & 31.8 & 198 \\
                                    & 256 &78.5 & 64.4 & 47.8 & 32.4& 163 \\

  \bottomrule
  \end{tabular}%
  \caption{Results on the CIFAR10 dataset on standard and  provably certifiable accuracies for different values of perturbations $\varepsilon$ on CPL (ours) models with various batch sizes. The average time per epoch in seconds is also reported in the last column. All the reported networks use Last Layer Normalization.}
  \label{table:c10-comp-bs}%
\end{table*}%

\begin{table*}[h]
  \centering
  \sisetup{%
    table-align-uncertainty=true,
    separate-uncertainty=true,
    detect-weight=true,
    detect-inline-weight=math
  }
  \begin{tabular}
  {
    l
    S[table-format=2.2]
    S[table-format=2.2]
    S[table-format=2.2]
    S[table-format=2.2]
    S[table-format=2.2]
    S[table-format=2.2]
  }
  \toprule
   &\multicolumn{1}{c}{\textbf{Batch size }}& \multicolumn{1}{c}{\textbf{Clean Accuracy}} & \multicolumn{3}{c}{\textbf{Provable Accuracy ($\varepsilon $)}} &  \multicolumn{1}{c}{\textbf{Time per epoch (s)}} 
    \\
    \cmidrule{4-6}
    & \multicolumn{1}{c}{ } & \multicolumn{1}{c}{ } &\multicolumn{1}{c}{36/255} & \multicolumn{1}{c}{72/255} &  \multicolumn{1}{c}{108/255} & \multicolumn{1}{c}{\textbf{}} 
    \\
  \midrule
  \multirow{3}{*}{\textbf{CPL-S}} & 64 & 45,6 & 30,8 & 19,3 & 11,2 & 47 \\
                                    & 128 & 44,9 & 30,7 & 19,2 & 11,0 & 31\\
                                    & 256 & 44,0 & 29,9 & 19,1 & 10,9 & 23\\
  \midrule
  \multirow{3}{*}{\textbf{CPL-M}} & 64 & 46.6 & 31,6 & 19,6 & 11,6 & 78 \\
                                    & 128 & 46.3 & 31,1 & 19,7 & 11,5 & 55 \\
                                    & 256 & 45.6 & 31,1 & 19,3 & 11,3 & 41 \\

  \midrule

  \multirow{3}{*}{\textbf{CPL-L}} & 64 & 48.1 & 32,7 & 20,3 & 11,7 & 163 \\ 
                                    & 128 & 47,4 & 32,3 & 20,0 & 11,8 & 116 \\ 
                                    & 256 & 46,8 & 31,8 & 20,1 & 11,7 & 95 \\ 
  \midrule
  \multirow{3}{*}{\textbf{CPL-XL}} & 64 & 49,0 & 33,7 & 21,1 & 12,0 & 293 \\
                                    & 128 & 48,0 & 33,7 & 21,0 & 12,1 & 209 \\
                                    & 256 &47,8 & 33,4 & 20,9 & 12,6 & 164 \\

  \bottomrule
  \end{tabular}%
  \caption{Results on the CIFAR100 dataset on standard and  provably certifiable accuracies for different values of perturbations $\varepsilon$ on CPL (ours) models with various batch sizes. The average time per epoch in seconds is also reported in the last column. All the reported networks use Last Layer Normalization.}
  \label{table:c100-comp-bs}%
\end{table*}%

In Tables~\ref{table:c10-comp-bs} and~\ref{table:c100-comp-bs}, we tried three different batch sizes (64, 128 and 256) for training our networks on CIFAR10 and CIFAR100 datasets, we remark a gain in standard accuracy in reducing the batch size for all settings. As the perturbation becomes larger, the gain in accuracy is reduced and can even in some cases we may loose some points in robustness.

\newpage
\subsection{Effect of the Margin Parameter}

\begin{figure}[h]
    \centering
    \begin{tabular}{cc}
    \includegraphics[width=0.49\textwidth]{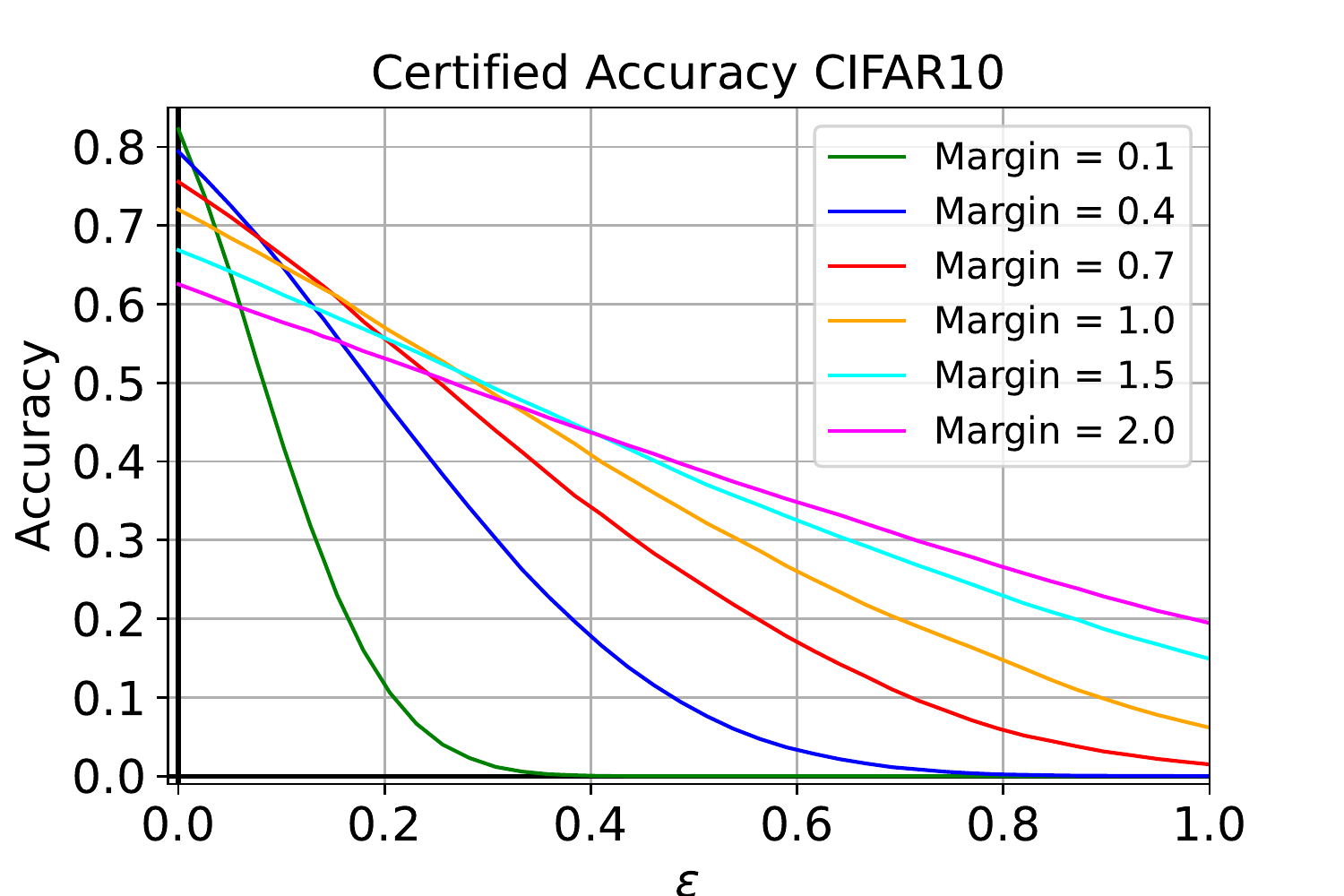}&\includegraphics[width=0.49\textwidth]{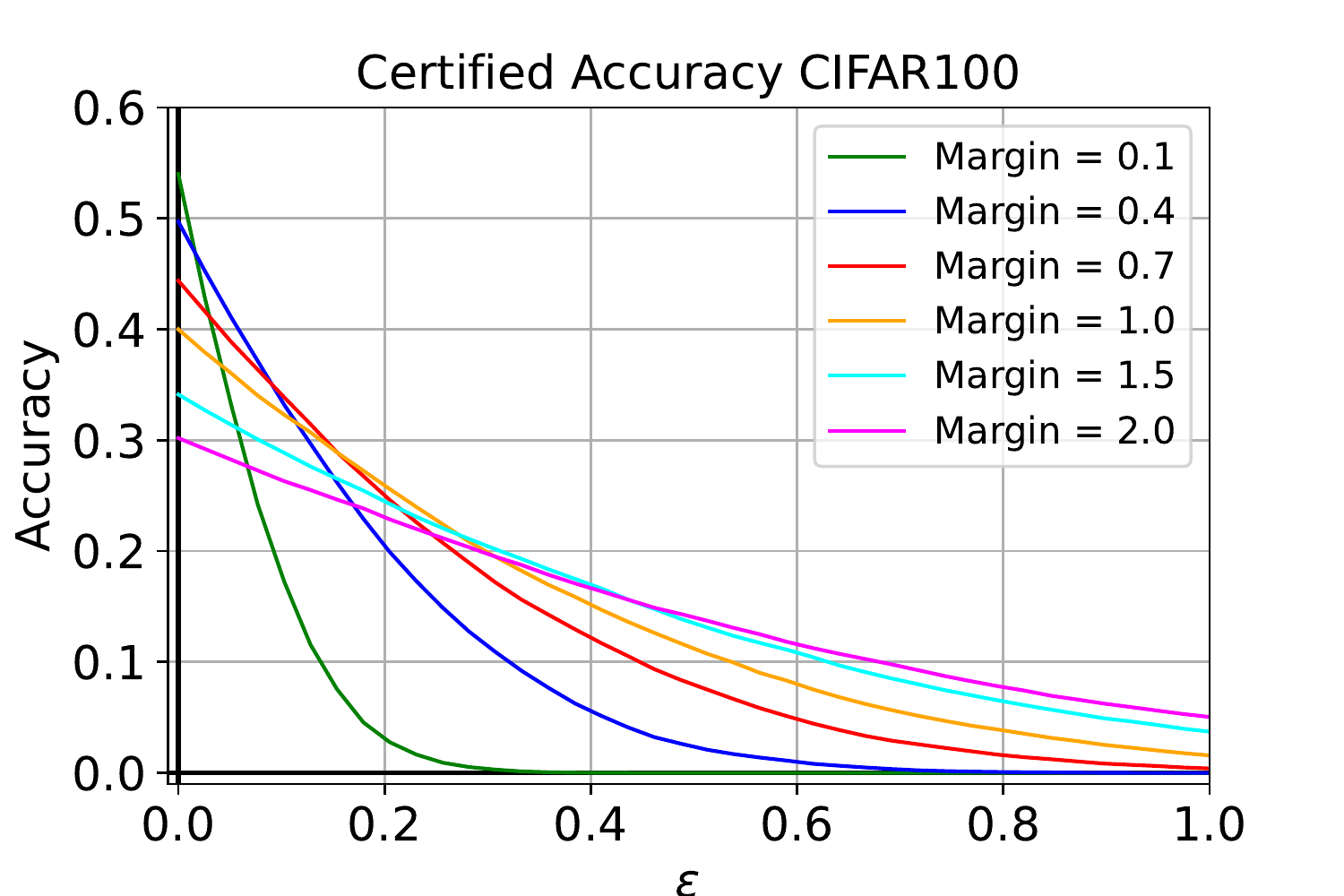}
    \end{tabular}
    \caption{Certifiably robust accuracy in function of the perturbation $\varepsilon$ for our CPL-S  network with different margin parameters on CIFAR10 and CIFAR100 datasets.}
    \label{fig:cert-acc-margin}
\end{figure}

\begin{figure}[h]
    \centering
    \begin{tabular}{cc}
    \includegraphics[width=0.49\textwidth]{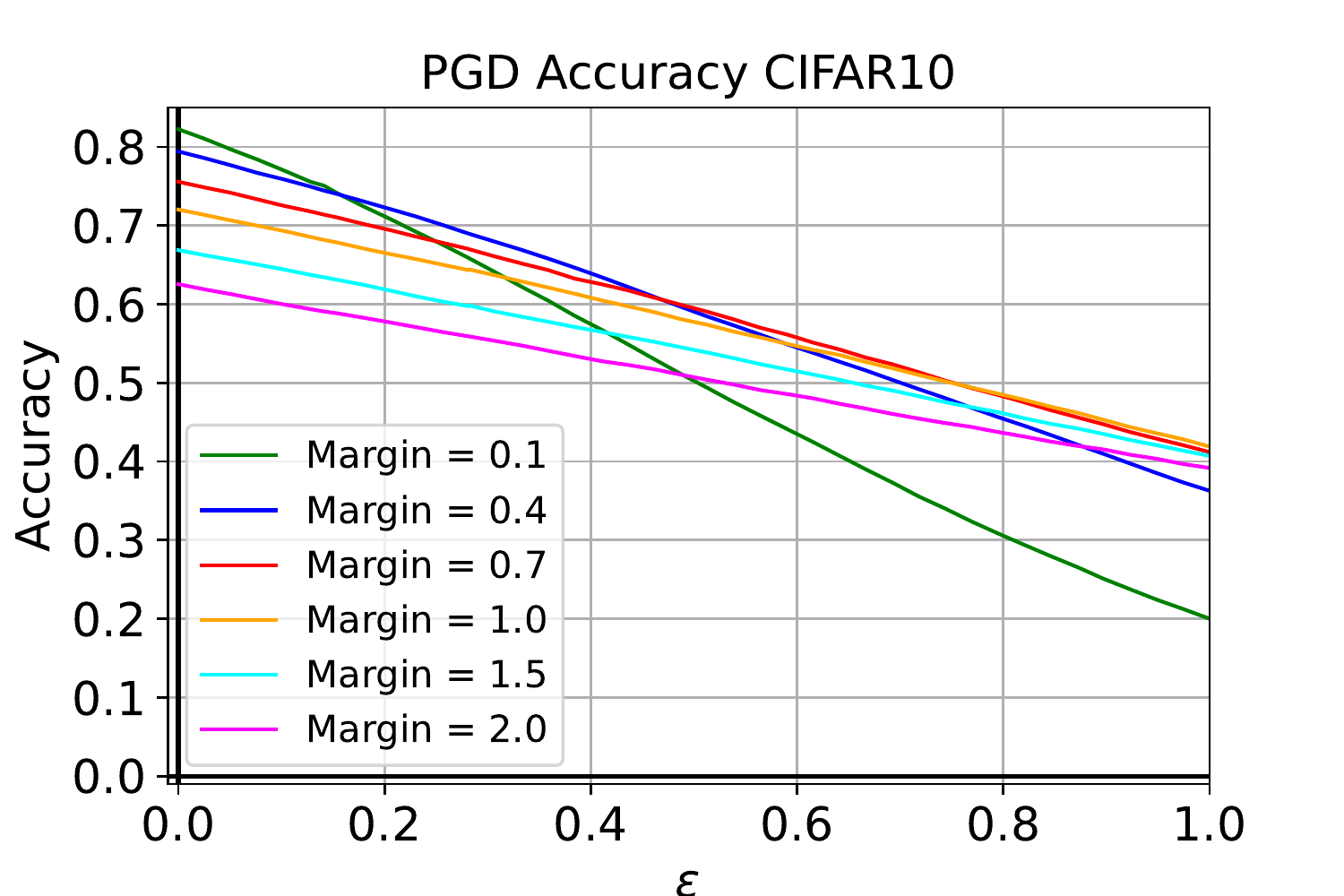}&\includegraphics[width=0.49\textwidth]{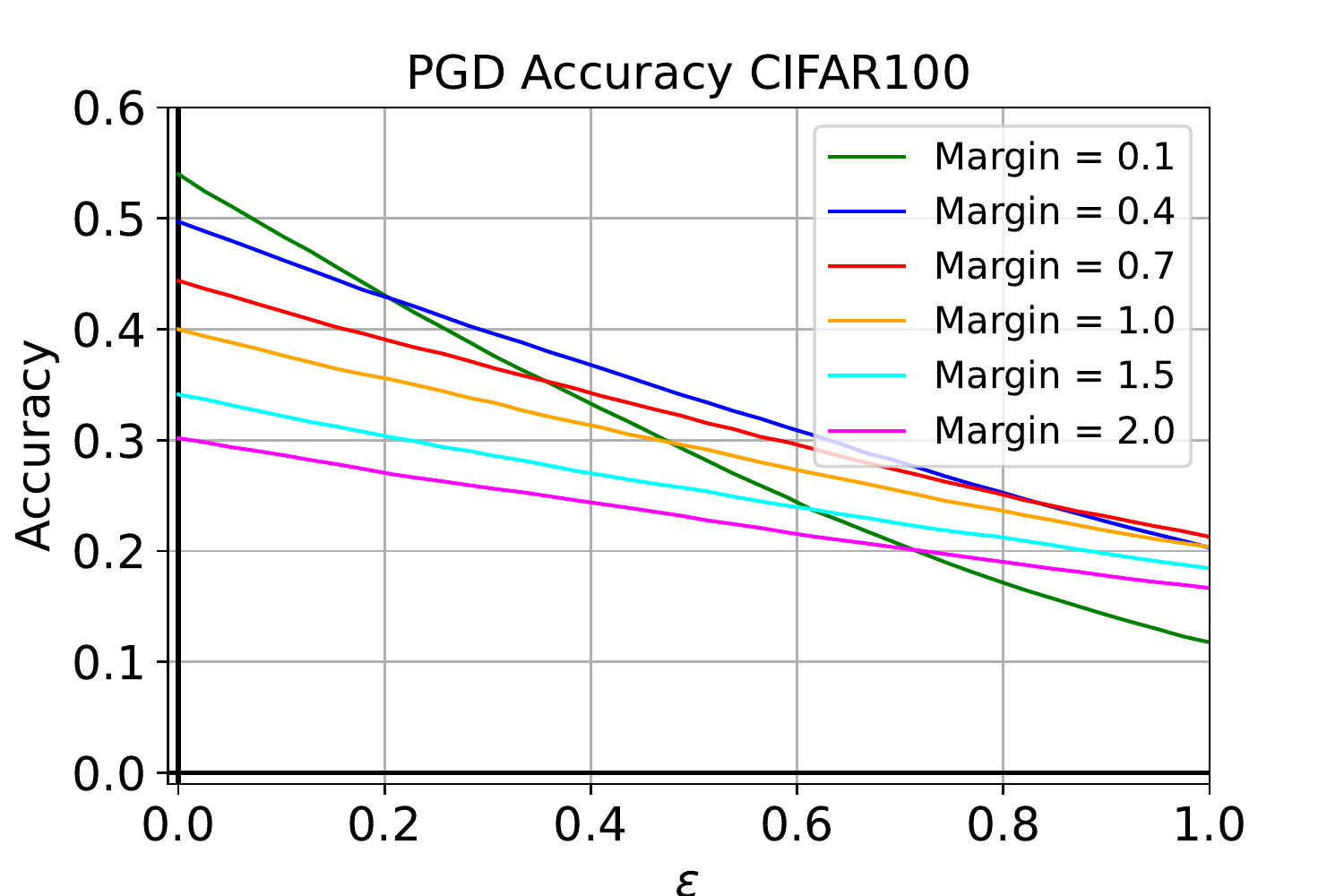}
    \end{tabular}
    \caption{Certifiably robust accuracy in function of the perturbation $\varepsilon$ for our CPL-S  network with different margin parameters on CIFAR10 and CIFAR100 datasets.}
    \label{fig:pgd-acc-margin}
\end{figure}

In these experiments we varied the margin parameter in the margin loss in Figures~\ref{fig:cert-acc-margin} and~\ref{fig:pgd-acc-margin}. It clearly exhibits a tradeoff between standard and robust accuracy. When the margin is large, the standard accuracy is low, but the level of robustness remain high even for ``large'' perturbations. On the opposite, when the margin is small, we get a high standard accuracy but we are unable to keep a good robustness level as the perturbation increases. It is verified both on certified and empirical robustness.

\end{document}